\documentclass[letterpaper, 10 pt, conference]{ieeeconf}  

\IEEEoverridecommandlockouts                              

\usepackage{graphicx} 
\usepackage{epstopdf} 
\usepackage{times} 
\usepackage{amsmath} 
\usepackage{amssymb}  
\makeatletter
\let\MYcaption\@makecaption
\makeatother
\usepackage{subcaption}
\makeatletter
\let\@makecaption\MYcaption
\makeatother
\usepackage{siunitx}
\usepackage{bm}
\usepackage[dvipsnames]{xcolor}
\usepackage{balance}
\usepackage{tikz}
\usepackage{csquotes}
\usepackage{cite}

\usepackage{titlesec}
\titleformat{\subsubsection}[runin]
  {\normalfont\normalsize\itshape}
  {\Alph{subsection}.\arabic{subsubsection})}
  {1em}{}[:]
\titlespacing*{\subsubsection}{0pt}{\parsep}{1em}                        
\title{\LARGE \bf
Safe Ergodic Control for Multi-Robot Systems via Quadratic Programming
}

\author{Yo Toyomoto$^{1}$, Mohamed Elobaid$^{1}$, Bryce L. Ferguson$^{2}$, Alberto Quattrini Li$^{3}$ and Shinkyu Park$^{1}$
\thanks{$^\star$This work was supported by funding from King Abdullah University of Science and Technology (KAUST) and the Dartmouth Initiative for Middle East Exchange (DIMEX).}
\thanks{$^{1}$Yo Toyomoto, Mohamed Elobaid and Shinkyu Park with Electrical and Computer Engineering, King Abdullah University of Science and Technology (KAUST), Thuwal 23955, Saudi Arabia. {\tt \small \{firstName.lastName\}@kaust.edu.sa}}%
\thanks{$^{2}$Bryce L. Ferguson is with the Thayer School of Engineering at Dartmouth College, Hanover, NH, USA. {\tt\small bryce.l.ferguson@dartmouth.edu}}%
\thanks{$^{3}$Alberto Quattrini Li is with the Department of Computer Science, Dartmouth College, Hanover, NH, USA. {\tt\small alberto.quattrini.li@dartmouth.edu}}%
}%

\newtheorem{theorem}{Theorem}

\newtheorem{remark}{Remark}

\begin{document}

\maketitle
\thispagestyle{empty}
\pagestyle{empty}

\begin{abstract}
Ergodic control drives robots to spend time in each region in proportion to a spatial distribution of interest, making it well suited for dense spatiotemporal environmental monitoring. Existing safe ergodic controllers rely on offline trajectory optimization or hierarchical architectures, which limit real-time applicability and decouple the ergodicity objective from the safety constraint. This paper presents a quadratic programming (QP)-based controller that treats ergodicity and safety jointly. We first introduce a Gaussian-kernel ergodic metric that, unlike the classical indicator-based metric, is time differentiable. This allows the exponential decay of the metric to be imposed as a time-varying control barrier function (CBF) constraint, relaxed by a slack variable, alongside hard CBF constraints for region containment and inter-robot collision avoidance. We establish that the resulting QP remains feasible from any safe initial configuration and that, up to the slack term, the ergodic metric decays exponentially. Simulations show improved performance over baseline methods, and experiments with three aerial vehicles validate the work on real hardware.
\end{abstract}


\section{Introduction}
Employing multiple mobile sensors, such as underwater robots and aerial vehicles with onboard sensors, yields far denser spatiotemporal data over a vast field than manual sampling \cite{robot-as-sensors}. A key challenge, however, lies in operating these robots efficiently; in particular, adapting their motion to an area of interest is critical to monitoring performance (e.g., through assigning a probability density function as in search and rescue scenarios \cite{sar-fixed-importance}, or mimicking heat maps corresponding to industrial inspection). 
In this context, coverage path planning has been widely studied, seeking the shortest path that covers the field \cite{cpp-survey, asvs-cpp}. Coverage control has also been applied to distribute multiple robots such that their stationary spatial distribution matches the area of interest \cite{coverage-control, coverage-review}. However, neither approach directly optimizes the time history of the robots' positions to match the spatial distribution required for monitoring the area.

Ergodic control instead matches the robots' trajectories to the target spatial distribution.
Mathew and Mezi\'{c} proposed in \cite{ergodic-metric} an ergodic metric that quantifies the discrepancy between the fraction of time the robots spend in each region and the target distribution. Because this metric is difficult to use directly in controller design, they proposed the spectral multiscale coverage (SMC) method, which approximates both distributions using Fourier series and derives a closed-form control input that minimizes the discrepancy between their Fourier coefficients. Fourier series-based ergodic control has since been widely studied \cite{ergodic-RH, ergodic-continuous, ergodic-SE2, ergodic-continuous-journal, ergodic-energy, ergodic-decentralized, ergodic-stein, ergodic-stochastic}, and alternative approximations of the ergodic metric, e.g., using heat equations~\cite{ergodic-HEDAC}, kernel functions~\cite{ergodic-kernel}, sample-based kernel embedding~\cite{ergodic-kernel-general} and optimization methods, e.g., learning methods~\cite{ergodic-learning}, and flow matching methods~\cite{ergodic-flow-matching}, have been explored. However, none of these methods guarantee safety, such as collision avoidance between robots. Several studies incorporate safety constraints in ergodic trajectory optimization~\cite{ergodic-safety-critical, ergodic-time-opti, ergodic-connectivity}, but, to the best of our knowledge, all of them are executed offline or rely on a long optimization horizon, which makes them difficult to adapt online. Hierarchical approaches that run SMC in an upper layer and a control barrier function (CBF)-based safety filter in a lower layer have also been proposed~\cite{ergodic-hierarchical, ergodic-hierarchical-2}. This hierarchical structure, however, cannot guarantee decay of the ergodic metric as the safety filter may degrade ergodicity convergence.

\begin{figure}[t]
    \centering
    \includegraphics[width=1\linewidth]{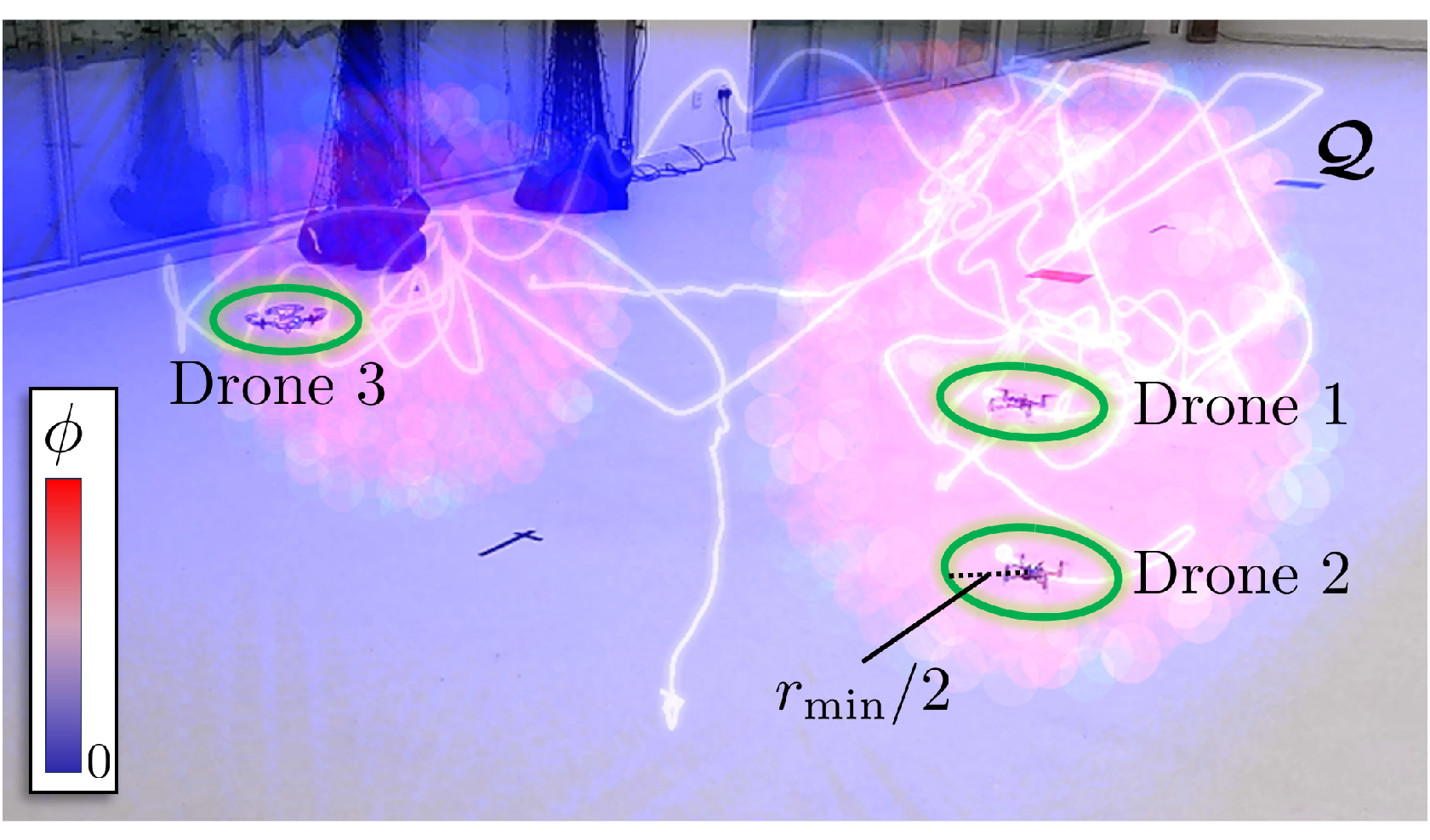}
    \caption{Safe ergodic control of multiple robots. Three drones move within a region $\mathcal Q$ so that the time spent along their trajectories (white curves) is distributed according to a desired distribution $\phi$ (color scale), while remaining inside $\mathcal Q$ and keeping a mutual distance of at least $r_{\min}$ (green circles of radius $r_{\min}/2$).}
    \label{fig:problem-ergodic}
\end{figure}

This paper presents a single-layer quadratic programming (QP)-based controller that decreases the ergodic metric while ensuring safety. To this end, we first define an ergodic metric using a Gaussian kernel which renders the metric time differentiable. This metric definition also differs from~\cite{ergodic-kernel} in that we apply the Gaussian filter to all terms of Equation (23) in~\cite{ergodic-kernel}, which more accurately captures the similarity of the trajectory to the target distribution. This property allows us to design the controller with a constraint-driven approach~\cite{robot-ecology}, which can be executed online in real time from the current state. Specifically, the decay rate of the ergodic metric is imposed as a constraint that is affine in the control input, structured as a time-varying CBF~\cite{t-CBF}, so that additional safety-critical constraints can be added as CBFs by construction.
Unlike hierarchical approaches~\cite{ergodic-hierarchical, ergodic-hierarchical-2}, the control input is optimized by considering the ergodic-metric decay and the safety constraints simultaneously.
Simulations compare the decay of the ergodic metric under the proposed controller against the SMC and hierarchical SMC-CBF methods, and experiments with multiple Crazyflies, as shown in Figure~\ref{fig:problem-ergodic}, validate the controller design. 

The rest of the paper is organized as follows. Notations are introduced below. Section~\ref{sec:ergodic_metric_formulation} introduces a novel ergodic metric using a Gaussian kernel, and the control problem is formulated. Section~\ref{sec:ergodic-motion-control} presents a controller design based on a CBF-based QP approach. Section~\ref{sec:validation} validates the controller through simulations across various scenarios against baseline methods, and further experimental results with Crazyflies are presented. Section~\ref{sec:conclusion} concludes the paper and discusses directions for future work.

\subsection{Notation}
Bold lowercase letters denote vectors, capital letters denote matrices. $\operatorname{diag}(a,b,c)$ denotes the diagonal matrix with entries $a,b,c$, and $A^{\circ k}$ its element-wise $k$-th power (likewise for vectors). $\|\cdot\|_k$ is the $k$-norm on $\mathbb{R}^d$.
We write $f(x;\theta)$ when $\theta$ is a parameter held fixed. A continuous, strictly increasing function $\xi:\mathbb{R}\to\mathbb{R}$ with $\xi(0)=0$ and $\xi(r)\to\pm\infty$ as $r\to\pm\infty$ is an \emph{extended class-$\mathcal{K}_\infty$} function.

\section{Problem Formulation with a Gaussian-Kernel Ergodic Metric}\label{sec:ergodic_metric_formulation}
We consider $m$ robots monitoring a closed subset $\mathcal{Q} \subset \mathbb{R}^3$, as illustrated in Fig.~\ref{fig:problem-ergodic}. The position of robot $j$ at time $t$ is denoted as $\bm{p}_j(t) \in \mathcal{Q}$, and its dynamics are those of a single integrator:
\begin{equation}
    \dot{\bm{p}}_j(t) = \bm{u}_j(t).
    \label{eq:robot-dynamics}
\end{equation}
The positions and control inputs are denoted by $\bm{p}(t) = (\bm{p}_1(t), \dots, \bm{p}_m(t))$ and $\bm{u}(t) = (\bm{u}_1(t), \dots, \bm{u}_m(t))$, respectively.

We define a target coverage distribution as a map
$\phi : \mathcal{Q} \to \mathbb{R}_{\ge0}$ with
$\int_{\mathcal{Q}} \phi(\bm{q}) \, d\bm{q} = 1$,
and extend it by zero to $\mathbb{R}^3 \setminus \mathcal{Q}$.
It specifies the desired distribution of the robots' time over $\mathcal Q$, i.e., the fraction of time spent in a region should be proportional to the integral of $\phi$ over that region.
Our goal is to drive the robots so that the distribution of their time coincides with $\phi$, that is, to make their trajectories \textit{ergodic} with respect to $\phi$~\cite{ergodic-theory-book}.

Because the robots are moving, the time spent \emph{at} a point $\bm q$ is not well defined. We therefore measure the time spent \emph{near} $\bm q$ through the Gaussian kernel
\begin{equation}
    \kappa(\bm{y}, \bm{q}; \sigma) = e^{-\frac{\|\bm{y-q}\|^2}{2\sigma^2}},
    \label{eq:gaussian kernel}
\end{equation}
which acts as a smooth indicator of a neighborhood of $\bm q$ of size $\sigma$. All quantities below depend on the robots' trajectories over $[0, t]$; for concise presentation, we omit this dependence from the notation.
Let
\begin{align}
    \bar\kappa(t, \bm q; \sigma) = \frac{1}{m}\sum_{j=1}^m \kappa(\bm p_j(t), \bm q; \sigma)
    \label{eq:kernel-average}
\end{align}
denote the kernel value averaged over the robots at time $t$. Its running time average,
\begin{equation}
    d(t,\bm{q},\sigma) = \frac{1}{t}\int_0^t \bar\kappa(\tau, \bm q; \sigma)\, d\tau,
    \qquad t > 0
    \label{eq:traj-ratio}
\end{equation}
is the fraction of time the robots have spent near $\bm q$. The corresponding target fraction is the target distribution filtered by the same kernel,
\begin{equation}
    \bar{\phi} (\bm{q},\sigma) = \int_{\mathcal{Q}} \kappa(\bm{y},\bm{q};\sigma)\, \phi(\bm{y}) \, d\bm{y}.
    \label{eq:target-ratio}
\end{equation}
By the ergodic theorem \cite{ergodic-theory-book}, the trajectories are ergodic with respect to $\phi$ if the time average of every integrable function converges to its $\phi$-weighted spatial average. Since $0<\kappa(\cdot, \bm q; \sigma)\leq1$ and hence (\ref{eq:target-ratio}) is well-defined,
this yields the control objective
\begin{equation}
    \lim_{t\rightarrow\infty} d(t,\bm{q}, \sigma) = \bar{\phi} (\bm{q}, \sigma) \quad \text{for } \bm q \in \mathcal Q \text{ and } \sigma > 0.
    \label{eq:def-ergodicity-math}
\end{equation}

To quantify the distance from ergodicity, we integrate the squared mismatch over $\mathcal Q$ and over a range of scales $[\sigma_{\min}, \sigma_{\max}]$:
\begin{equation}
 \mathcal{E} (t) = \int_{\sigma_{\mathrm{min}}}^{\sigma_{\mathrm{max}}}\int_{\mathcal{Q}} \big( 
 d(t,\bm{q}, \sigma)
 -  \bar{\phi}(\bm{q}, \sigma) 
 \big)^2 \, d\bm{q} \, d\sigma, \ t > 0.
    \label{eq:smooth-ergodic}
\end{equation}
For each $t>0$, the integrand is smooth in $(\bm q, \sigma)$ on the compact set $\mathcal Q \times [\sigma_{\min}, \sigma_{\max}]$ and differentiable in $t$, so $\mathcal E$ is finite, non-negative, and differentiable in $t$. Moreover, $\mathcal E(t) = 0$ if and only if $d(t,\bm{q}, \sigma) =  \bar{\phi}(\bm{q}, \sigma)$ for all $\bm q$ and $\sigma$. Compared to the metric of \cite{ergodic-metric}, which uses an indicator function in place of $\kappa$, the Gaussian kernel is what makes $\mathcal E$ differentiable in time.

The control input enters $\mathcal E$ through its second time derivative, which is affine in $\bm u$. To see this, write $\kappa_j := \kappa(\bm p_j(t), \bm q; \sigma)$ and, here and in the remainder of this section, omit the arguments $(t, \bm q, \sigma)$ of $d$, $\bar\phi$, and the quantities derived from them. Differentiating \eqref{eq:traj-ratio} twice in $t$ gives
\begin{align*}
    \dot d &= -\frac{1}{t}\, d + \frac{1}{mt}\sum_{j=1}^m \kappa_j, \\
    \ddot d &= \underbrace{\frac{2}{t^2}\, d
        - \frac{2}{mt^2}\sum_{j=1}^m \kappa_j}_{=:\ \ddot d^{\,0}}
        \;+\; \sum_{j=1}^m \bm\rho_j^\top \bm u_j,
\end{align*}
where $\bm\rho_j := -\frac{\kappa_j}{m\sigma^2 t}\,\big(\bm p_j(t) - \bm q\big)$ and $\ddot d^{\,0}$ collects the terms that do not multiply $\bm u$.
Substituting into $\ddot{\mathcal E} = \int\!\!\int \big[ 2\dot d^{\,2}
+ 2(d - \bar\phi)\,\ddot d \big]\, d\bm q\, d\sigma$ yields
\begin{equation}\label{eq:affine-form-of-Es}
    \ddot{\mathcal E}(t, \bm u) = a(t) + \sum_{j=1}^m \bm b_j(t)^\top \bm u_j ,
\end{equation}
where $a(t) \in \mathbb R$ and $\bm b_j(t) \in \mathbb R^3$ are given by
\begin{align*}
    a(t) &= \int_{\sigma_{\min}}^{\sigma_{\max}}\!\!\int_{\mathcal Q} \Big[ 2\dot d^{\,2} + 2\big(d - \bar\phi\big)\,\ddot d^{\,0} \Big] \, d\bm q\, d\sigma, \\
    \bm b_j(t) &= \int_{\sigma_{\min}}^{\sigma_{\max}}\!\!\int_{\mathcal Q} 2\big(d - \bar\phi\big)\, \bm\rho_j \, d\bm q\, d\sigma .
\end{align*}

Since $\mathcal{E}$ depends on the entire trajectory history and the control input enters only through $\ddot{\mathcal{E}}$, the ergodic metric cannot be driven to zero instantaneously. Therefore, we approach (\ref{eq:def-ergodicity-math}) by enforcing a prescribed decay rate $\gamma$ on $\mathcal{E}$ at each instant, while maintaining a prescribed safety distance between robots and keeping them in $\mathcal{Q}$.

\section{Safe Ergodic Motion Control via Quadratic Programming}\label{sec:ergodic-motion-control}
For $t\ge t_0 > 0$ and a fixed decay rate $\gamma >0$, define
\begin{equation}
    h(t) := -\dot{\mathcal{E}}(t) - \gamma\mathcal{E}(t),
    \label{eq:def-h}
\end{equation}
so that $h(t) \geq 0$ enforces an exponential decay of $\mathcal{E}$ at rate $\gamma$. Let $\bm b(t) = (\bm b_1(t), \dots, \bm b_m(t))$ collect the vectors in \eqref{eq:affine-form-of-Es}. Then \eqref{eq:affine-form-of-Es} gives
\begin{align} \label{eq:h_dot}
    \dot h(t, \bm u) &= -a(t) - \gamma \mathcal{\dot E}(t) - \bm b(t)^\top \bm u,
\end{align}
which is affine in $\bm u$.

Following the CBF-based QP approach of \cite{t-CBF}, we treat $h$ as a time-varying control barrier function. Given an extended class-$\mathcal{K}_\infty$ function
$\alpha$, the \textit{relaxed} control barrier function $\dot{h} + \alpha(h) + \delta \geq 0$ reads, by \eqref{eq:h_dot},
\begin{align}
     \psi(t) - \bm b (t)^\top \bm u + \delta &\geq 0,
\end{align}
where $\psi(t) := -a(t) - \gamma \mathcal{\dot E} (t) + \alpha(h(t))$ and $\delta$ is a slack variable. In addition, the robots must remain in $\mathcal Q$ and avoid collisions with one another; we encode both requirements as CBFs. Assume $\mathcal Q$ is a box with center $\bm q_c$ and half-widths $\ell_1, \ell_2, \ell_3$. For robot~$j$, define
\begin{equation}\label{eq:boundary_cbf}
h_{\mathrm{b}, j} (\bm{p}_j)=  1 - \|W(\bm{p}_j - \bm{q}_\mathrm{c})\|_4^4,
\end{equation}
with $W = \operatorname{diag}(1/\ell_1, 1/\ell_2, 1/\ell_3)$. The $4$-norm is chosen as it smooths the vertices of the box while remaining a closer approximation to its shape than the $2$-norm. For robots $j$ and $k$, let $r_{\mathrm{min}} > 0$ be a minimum inter-robot distance and define
\begin{equation}\label{eq:collision_cbf}
    h_{\mathrm{c},jk}(\bm{p}_{j}, \bm{p}_{k}) = \|\bm{p}_{j} - \bm{p}_{k} \|_2^2 - r_{\mathrm{min}}^2.
\end{equation}
Both functions are smooth,
with gradients $\nabla_{\bm p_j}h_{\mathrm{b}, j} = -4W^{\circ4}(\bm{p}_j - \bm{q}_\mathrm{c})^{\circ3}$ and $\nabla_{\bm p_j} h_{\mathrm{c},jk} = 2(\bm{p}_j-\bm{p}_k)$, so that $\dot h_{\mathrm{b}, j} = \nabla_{\bm p_j}h_{\mathrm{b}, j}^\top \bm u_j$ and $\dot h_{\mathrm{c},jk} = 2 (\bm p_j - \bm p_k)^\top (\bm u_j - \bm u_k)$ are also affine in $\bm u$. 

We can now cast the control problem as the convex QP:
\begin{subequations}
\begin{align}
    \bm{u}^{\star}, \delta^{\star} &= \underset{\bm{u}, \delta}{\mathrm{arg min}} \|\bm{u}\|_2^2 + w|\delta|^2 \label{eq:qp-constraint-ergodic-region-obj}\\
    \mathrm{s.t.} \quad & \dot{h} + \alpha\left(h\right)  + \delta \geq 0,\label{eq:qp-constraint-ergodic-region-obj-const-erg}\\
    & \dot{h}_{\mathrm{b},j} + \beta_{\mathrm{b}}(h_{\mathrm{b},j}) \geq 0 \quad \forall j,\label{eq:qp-constraint-ergodic-region-obj-const-region} \\
    & \dot{h}_{\mathrm{c},jk} + \beta_{\mathrm{c}}(h_{\mathrm{c},jk})\geq 0 \quad \forall j < k,\label{eq:qp-constraint-ergodic-region-obj-const-col} \\ 
    & \delta \geq 0,\label{eq:qp-constraint-ergodic-region-obj-const-slack}
\end{align}
\label{eq:qp-constraint-ergodic-region}
\end{subequations}
where $w>0$ is a (large) weight and $\beta_{\mathrm{b}}, \ \beta_{\mathrm{c}}$ are extended class-$\mathcal{K}_\infty$ functions. Only the ergodicity constraint~\eqref{eq:qp-constraint-ergodic-region-obj-const-erg} is relaxed through the slack; the safety constraints \eqref{eq:qp-constraint-ergodic-region-obj-const-region}--\eqref{eq:qp-constraint-ergodic-region-obj-const-col} are hard. The QP thus prioritizes safety and enforces \enquote{\textit{practical ergodicity}}, in the sense made precise by the following result.
\begin{theorem}\label{theo:no_actuator_limits}
    Let $\alpha(h) = c_\alpha h$, $\beta_{\mathrm{b}}(h_{\mathrm{b},j}) = c_{b}h_{\mathrm{b},j}$, $\beta_{\mathrm{c}}(h_{\mathrm{c},jk}) = c_ch_{\mathrm{c},jk}$ with $c_\alpha, c_b, c_c > 0$, and $c_\alpha\neq\gamma$. Suppose the robots are safe at $t_0 > 0$, that is,
    \begin{align*}
    h_{\mathrm{b},j}\big(\bm{p}_j(t_0)\big) &\ge0 \ \ \forall j,
    \\
    h_{\mathrm{c},jk}\big(\bm{p}_j(t_0),\bm{p}_k(t_0)\big) &\ge0 \ \ \forall j<k .
\end{align*}
Then, for all $t \geq t_0$, the QP \eqref{eq:qp-constraint-ergodic-region} is feasible, the robots remain safe, and
\begin{multline}\label{eq:ergodic_convergence_rate}
    \mathcal{E}(t)\ \le\
    \Big(\mathcal{E}(t_0)+\tfrac{\eta}{|\gamma-c_\alpha|}\Big)
    e^{-\min\{\gamma,\,c_\alpha\} (t - t_0)}
    \\ + \int_{t_0}^{t}e^{-\gamma(t-\tau)}\delta_c(\tau)\,d\tau,
\end{multline}
where $\eta := \max\{0, -h(t_0)\}$ is the initial violation of the decay
condition $h \ge 0$, and
\begin{equation*}
    \delta_{\mathrm c}(t) := \int_{t_0}^{t} e^{-c_\alpha (t - \tau)}\,
    \delta^\star(\tau)\, d\tau
\end{equation*}
is the accumulated slack, with $\delta^\star$ the optimal slack of
\eqref{eq:qp-constraint-ergodic-region}.
\end{theorem}

The proof of Theorem~\ref{theo:no_actuator_limits}, given in the Appendix, is constructive. It provides explicit expressions for the optimal control input, slack, and multipliers. Further, equation~\eqref{eq:ergodic_convergence_rate} states that, up to the slack term, the ergodic metric decays exponentially at rate $\min\{\gamma,c_\alpha\}$ from any safe initial condition. Moreover, the decay condition $h(t_0) \geq 0$ is not required at the startup; any violation enters the bound only through $\eta$. The proof also shows that $\bm u^\star$ and $\delta^\star$ remain well defined when $\bm{b}=0$, i.e., when the input has no instantaneous effect on $\ddot{\mathcal E}$. In that case the slack absorbs the ergodicity constraint and the input reduces to purely safety-driven motion.

\begin{remark}\label{rem:input-constraint}
In practice, actuator limits bound each robot's achievable velocity, as in the experiments of Section~\ref{sec:validation}. We therefore add to \eqref{eq:qp-constraint-ergodic-region} the input constraint
\begin{equation}\label{eq:qp-constraint-ergodic-region-obj-const-input}
    \|\bm u_j\|_{\infty} \leq u_{\max} \quad \forall j.
\end{equation}
This constraint preserves convexity of the QP, but the feasibility guarantee of Theorem~\ref{theo:no_actuator_limits} may be lost, since the hard safety constraints may require inputs exceeding $u_{\max}$. Whenever the QP remains feasible, the bound  \eqref{eq:ergodic_convergence_rate} still holds, but the input limit typically forces a larger slack $\delta^\star$ and hence a weaker bound.
\end{remark}

\section{Validation}\label{sec:validation}

\subsection{Simulation Setting}
We consider three simulation scenarios.\footnote{Additional simulation scenarios are provided in the supplemental video.} First, a single robot under a uniform target distribution is compared with the SMC method of \cite{ergodic-metric} to confirm that the proposed controller achieves comparable ergodicity. Second, multiple robots under two-peak are compared with the hierarchical SMC--CBF approach, showing that the proposed controller attains better ergodicity while guaranteeing safety. Third, a single robot under a two-peak distribution with increasing distance between the peaks is simulated to demonstrate that the controller reaches distant high-importance regions despite the potentially myopic nature of QP-based methods.

In all simulations, $\mathcal{Q}$ is a cube with side length $\SI{10}{m}$ discretized with a grid spacing of $\SI{0.1}{m}$ on which the target distribution and ergodic metric are evaluated. The maximum velocity of each robot is $u_{\max} = \SI{1.0}{m/s}$; the remaining controller parameters are listed in Table~\ref{tab:params-const-erg}.
The range $[\sigma_{\min}, \sigma_{\max}]$ for $\sigma$ is discretized with the same $\SI{0.1}{m}$ step as the grid, with $\sigma_{\min} = \SI{0.1}{m}$ equal to the grid spacing. Performance is sensitive to $\sigma_{\max}$, particularly when it is too small,
so its value is chosen by trading off computational load against the diminishing performance gain from larger values.
The decay rate $\gamma$ is chosen heuristically: a large $\gamma$ makes the condition $h(t) \geq 0$ hard to satisfy at all times, whereas a smaller $\gamma$ prescribes slow decay.
We start from a large $\gamma$ and reduce it, examining the decay of the ergodic metric over the first $\SI{20}{s}$.

The robot positions and $\bar \kappa$ are sampled every $\SI{0.01}{s}$ and accumulated into $d(t,\bm{q}, \sigma)$, with the metric initialized at $t_0=\SI{0.01}{s}$ from $\bm{p}(t_0)$. Since evaluating the ergodic metric and its derivatives is comparatively computationally expensive, they are computed every $\SI{0.2}{s}$, and each evaluation triggers one QP solve.
The QP is solved using the Python \texttt{osqp} package~\cite{osqp} with 5 polishing iterations, absolute and relative tolerances of $10^{-3}$, and at most $10^6$ iterations. If the solver reports primal infeasibility, the problem is re-solved without the input constraint \eqref{eq:qp-constraint-ergodic-region-obj-const-input} (cf.\ Remark~\ref{rem:input-constraint}). In our simulations, the solver rarely reaches the iteration limit and \textit{never} reported primal infeasibility.

\begin{table}[bht]
    \centering
    \begin{tabular}{c c | c c}
        \hline
        Parameter & Value &
        Parameter & Value \\
        \hline
        $\sigma_{\min}$ & $0.1$ &
        $\sigma_{\max}$ & $2.0$ \\ 
        $\alpha(\theta)$ & $\theta$ &
        $\gamma$ & $0.1$\\
        $w$ & $10^{12}u_{\max}$ &
        $\beta_{\mathrm{b}}(\theta)$ & $\theta$ \\
        $\beta_{\mathrm{c}}(\theta)$ & $0.25\theta$&
        $r_{\min}$ & $0.3$\\
        \hline
    \end{tabular}
    \caption{Parameters used in the simulations.}
    \label{tab:params-const-erg}
\end{table}

\subsection{Simulation Results}
\subsubsection{First scenario}
We compare the proposed QP controller with the SMC method of \cite{ergodic-metric}.
To match the input constraint $\|\bm u_j\|_{\infty} \leq u_{\max}$ of the proposed method, the SMC input is saturated as
\begin{equation}
    \bm u_j= - u_{\max} \frac{\bm \zeta_j}{\|\bm \zeta_j\|_{\infty}},
    \label{eq:SMC}
\end{equation}
where $\bm \zeta_j$ is the optimal velocity direction derived by SMC for robot~$j$. Due to computational limitations, the Fourier series is truncated at $10$ basis functions per dimension. 

\begin{figure}[tb]
    \centering
    \includegraphics[width=1.0\linewidth]{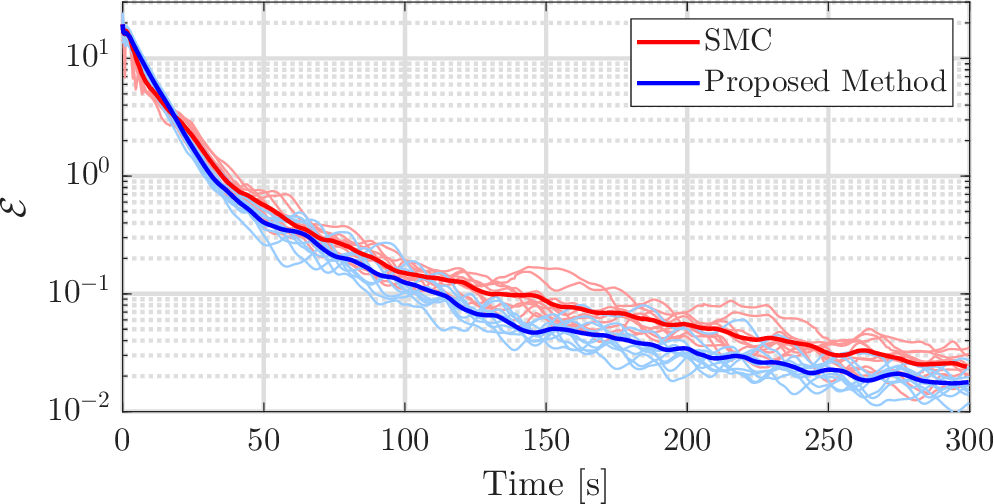}
    \caption{Time evolution of the ergodic metric $\mathcal E$ in \eqref{eq:smooth-ergodic} for a single robot under a uniform target distribution scenario with the proposed controller (blue) and SMC baseline (red). Thin lines show ten trials with random initial positions; bold lines show their averages.}
    \label{fig:ergodic-compare-uni}
\end{figure}

Figure~\ref{fig:ergodic-compare-uni} shows the time evolution of the ergodic metric (\ref{eq:smooth-ergodic}) for ten trials with random initial positions: thin blue lines for the proposed controller, thin red lines for the SMC baselines, and bold lines for the respective average.
Both methods reduce the metric to approximately $2.0\times10^{-2}$ by $t=\SI{300}{s}$, indicating comparable performance.

\begin{figure}
    \centering
    \includegraphics[width=1.0\linewidth]{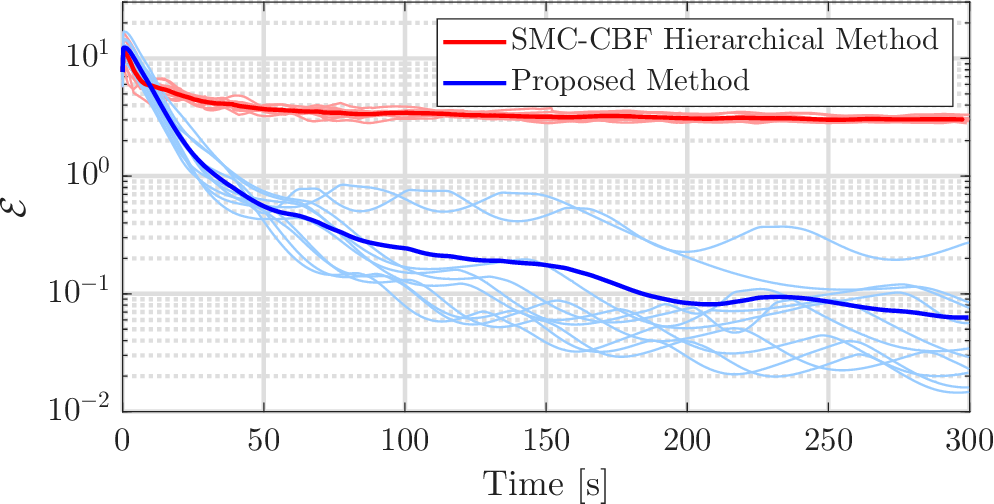}
    \caption{Time evolution of the ergodic metric $\mathcal E$ in \eqref{eq:smooth-ergodic} for three robots under a two-peak target distribution scenario with the proposed controller (blue) and hierarchical SMC--CBF baseline (red). Thin lines show ten trials with random initial positions; bold lines show their averages.}
    \label{fig:ergodic-compare-twin-multi}
\end{figure}

\subsubsection{Second Scenario} We next consider three robots under a two-peak target distribution, defined as a normalized sum of $n_\phi$ Gaussians,
\begin{equation}
\label{eq:multi-gauss-normalize}
 \phi(\bm{q})=\frac{\sum_{i=1}^{n_\phi} \frac{1}{n_\phi}\exp(-\|\bm{q}-\bm{q}_{\phi_i}\|_2^2/2\sigma_\phi^2)}{\int_\mathcal{Q}\sum_{i=1}^{n_\phi} \frac{1}{n_\phi}\exp(-\|\bm{q}-\bm{q}_{\phi_i}\|_2^2/2\sigma_\phi^2)\; d\bm{q}},
\end{equation}
with $n_\phi=2$ and $\sigma_{\phi} = 0.5$. The peak centers $\bm{q}_{\phi_i}$ are drawn at random once and kept fixed across trials.

We compare the proposed controller with the hierarchical SMC-CBF approach of \cite{ergodic-hierarchical} in which a lower-layer CBF-QP minimally modifies the nominal SMC input $\bm{u}_j^{\mathrm{nom}}$ of each robot to guarantee safety:
\begin{subequations}\label{eq:lower-safety-filter}
    \begin{align}
    \bm{u}_j^{\star} &= \underset{\bm{u}_j}{\mathrm{arg min}} \|\bm{u}_j - \bm{u}_j^{\mathrm{nom}}\|_2^2 \label{eq:lower-safety-filter-const-region} \\
    \mathrm{s.t.} \quad & \dot{h}_{\mathrm{b},j} + \beta_{\mathrm{b}}(h_{\mathrm{b},j}) \geq 0, \label{eq:lower-safety-filter-const-bou} \\
    & \nabla_{\bm p_j} h_{\mathrm{c},jk} \bm{u}_j + \frac{1}{2}\beta_{\mathrm{c}}(h_{\mathrm{c},jk})\geq 0 \quad \forall k \neq j.\label{eq:lower-safety-filter-const-col}
\end{align}
\end{subequations}
Constraint (\ref{eq:lower-safety-filter-const-col}) implements (\ref{eq:qp-constraint-ergodic-region-obj-const-col}) in distributed form, as in \cite{ergodic-hierarchical}. We keep this baseline distributed, since the nominal SMC command $\bm{u}_j^{\mathrm{nom}}$ is computed per robot; centralizing only the safety filter requires synchronization with these independent SMC outputs at the filter's own rate.
The lower-layer QP runs every $\SI{0.01}{s}$.

Figure~\ref{fig:ergodic-compare-twin-multi} shows the ergodic metric \eqref{eq:smooth-ergodic} for ten trials of each method from random initial positions: the blue thin lines for the proposed controller, thin red lines for the SMC-CBF baseline, and the bold lines for the averages. At $\SI{300}{s}$, the proposed controller reaches approximately $6\times10^{-2}$, whereas the baseline plateaus at approximately $3$, more than an order of magnitude higher. 

The gap is explained by the structure of the two controllers. In the hierarchical approach, the safety filter can redirect the input away from, or even opposite to, the SMC optimal direction whenever a collision-avoidance constraint becomes active. The proposed controller instead computes the input by jointly accounting for the decay of the ergodic metric and the safety constraints.

Despite the joint formulation, the discrete-time implementation causes minor, transient constraint violations: the box-boundary constraint $h_{\mathrm{b}, j}$ is violated by up to about $\SI{0.05}{m}$, and the minimum inter-robot distance reaches $\SI{0.25}{m}$. Since the CBF conditions render the respective safe set asymptotically stable \cite{robot-ecology}, the robots return to them in the following steps and continue exploring.
This is a known limitation of discretizing continuous-time CBF-based methods \cite{sample-CBF}, wherein the presented approach guarantees strict safety provided a fast enough sampling rate.

\subsubsection{Third Scenario} To examine the robot's behavior under increasingly distant peaks, we simulate a single robot with two-peak target distribution whose centers are placed symmetrically along the diagonal of the cube space $\mathcal Q$, 
$\bm{q}_{\phi_1} = \bm{q}_c + \ell_{\phi} \bm{1}$ and $\bm{q}_{\phi_2} = \bm{q}_c - \ell_{\phi} \bm{1}$, with $\ell_{\phi} = 1.0, 2.0, 3.0$.

\begin{figure}[tb]
    \centering
    \includegraphics[width=1.0\linewidth]{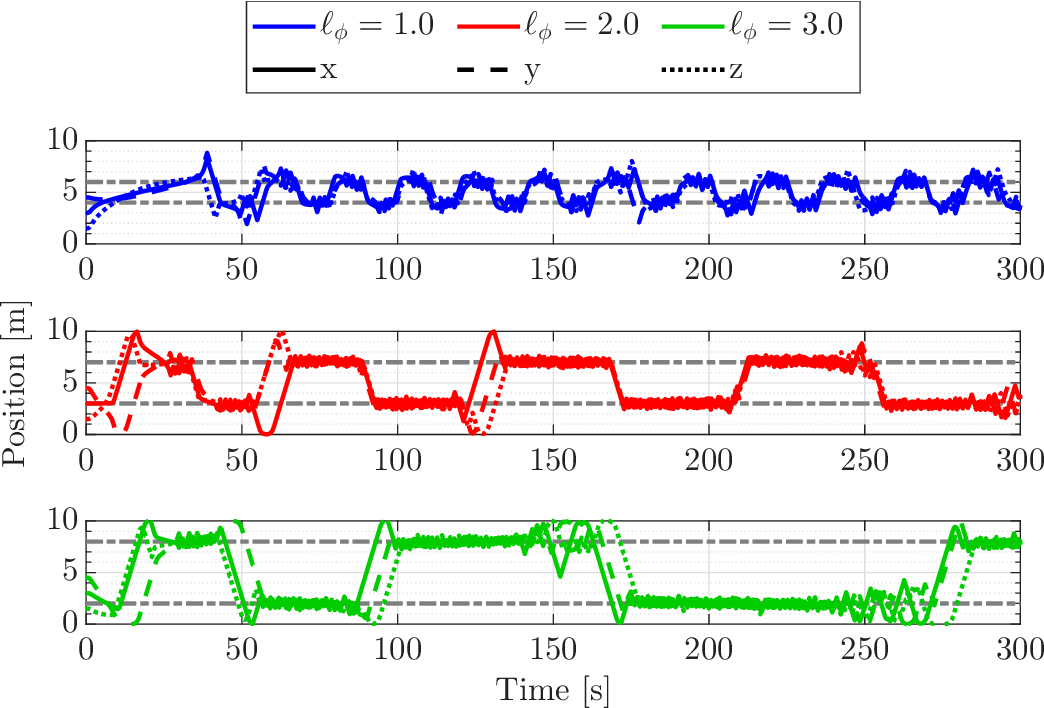}
    \caption{Robot position along $x$, $y$, and $z$ in the two-peak simulations with $\ell_\phi = 1.0$, $2.0$, and \SI{3.0}{m} (top to bottom). Gray dash-dotted lines mark the coordinates of the two peaks.}
    \label{fig:twin-position}
\end{figure}

\begin{figure}[tb]
    \centering
    \includegraphics[width=1.0\linewidth]{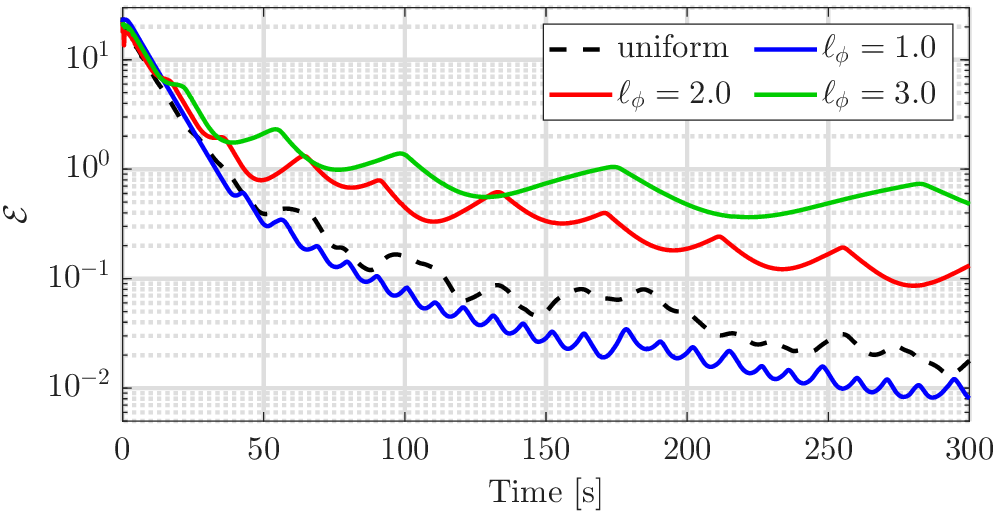}
    \caption{Time evolution of the ergodic metric $\mathcal E$ in \eqref{eq:smooth-ergodic} for a single robot under two-peak target distribution with separations $\ell_\phi \in \{1.0, 2.0, 3.0\}\,\si{m}$ (solid) and under the uniform target distribution of the first scenario (dashed).}
    \label{fig:ergodic-compare-twin}
\end{figure}

Figure~\ref{fig:twin-position} shows the robots' position for each case. The $x$-, $y$-, and $z$-coordinates frequently coincide, indicating robot's motion along the diagonal of the cube, and most of these segments occur when the robot is near one of the peaks, marked by the gray dash-dotted lines. Both peaks are visited in all three cases.
Figure~\ref{fig:ergodic-compare-twin} shows the corresponding ergodic metric (solid lines) together with the uniform-index case of the first scenario (dashed line) for reference. Initially the metric decreases at a rate similar to the uniform case; thereafter it oscillates, while its local minima continue to decrease.

Comparing the two figures explains the oscillations. While the robot stays near one peak, the time average $d$ there overshoots its target $\bar\phi$ and the deficit at the other peak grows, until no input near the current peak can decrease $\mathcal E$. The decay constraint \eqref{eq:qp-constraint-ergodic-region-obj-const-erg} then drives the robot toward the other peak; the metric rises during the transit and falls again once the robot arrives, and the cycle repeats.

Since the robot visits both peaks in every case and the local minima of the ergodic metric keep decreasing, we conclude that the proposed controller achieves ergodic behavior not only for a uniform target distribution but also when the mass of $\phi$ is concentrated in distant regions, despite the myopic nature of the QP.

\subsection{Drones Experiments}

\begin{figure*}[tb]
    \centering
    \begin{subfigure}[b]{0.32\linewidth}
        \includegraphics[width=\textwidth]{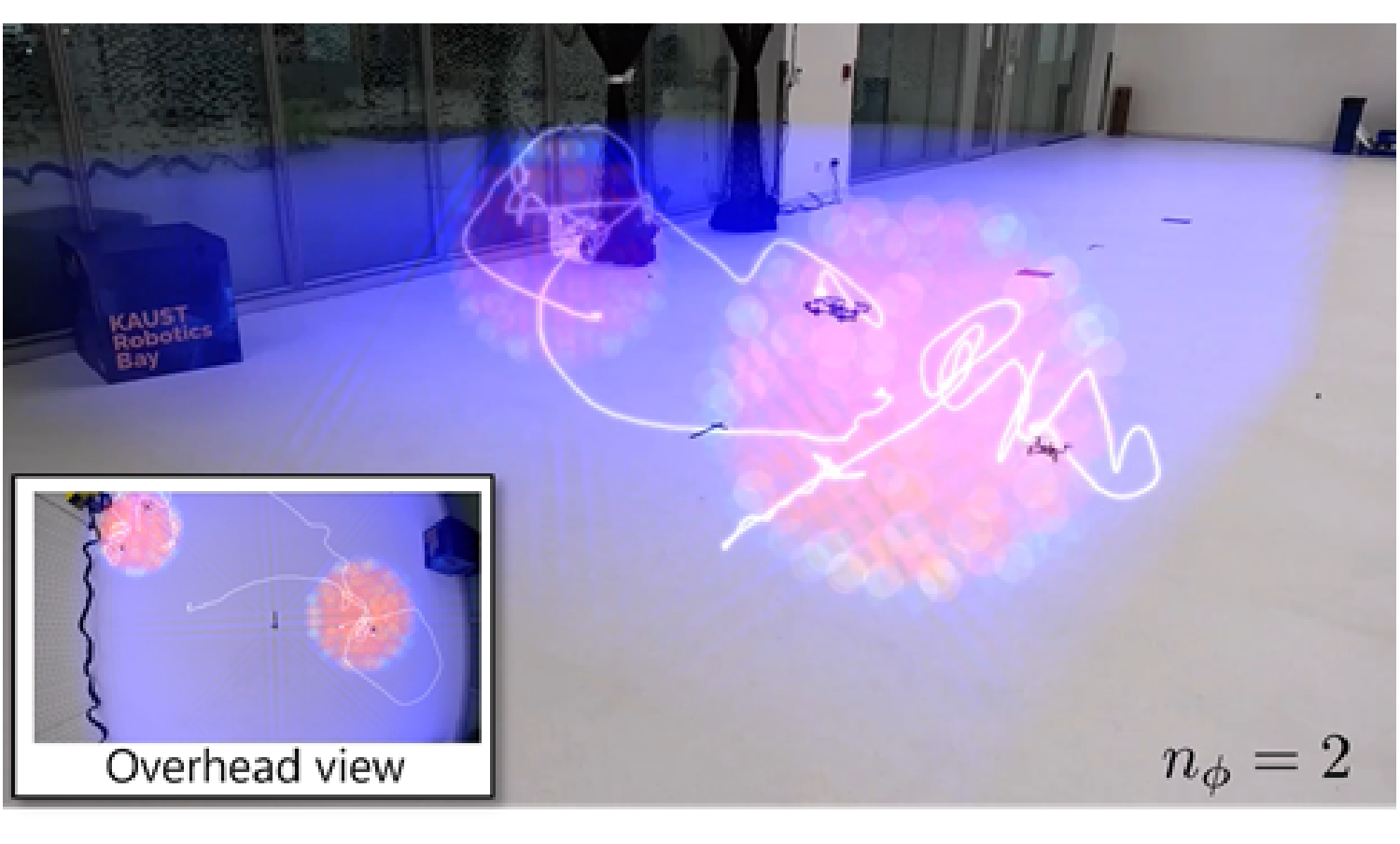}
    \end{subfigure}
    \hfill 
    \begin{subfigure}[b]{0.32\linewidth}
        \includegraphics[width=\textwidth]{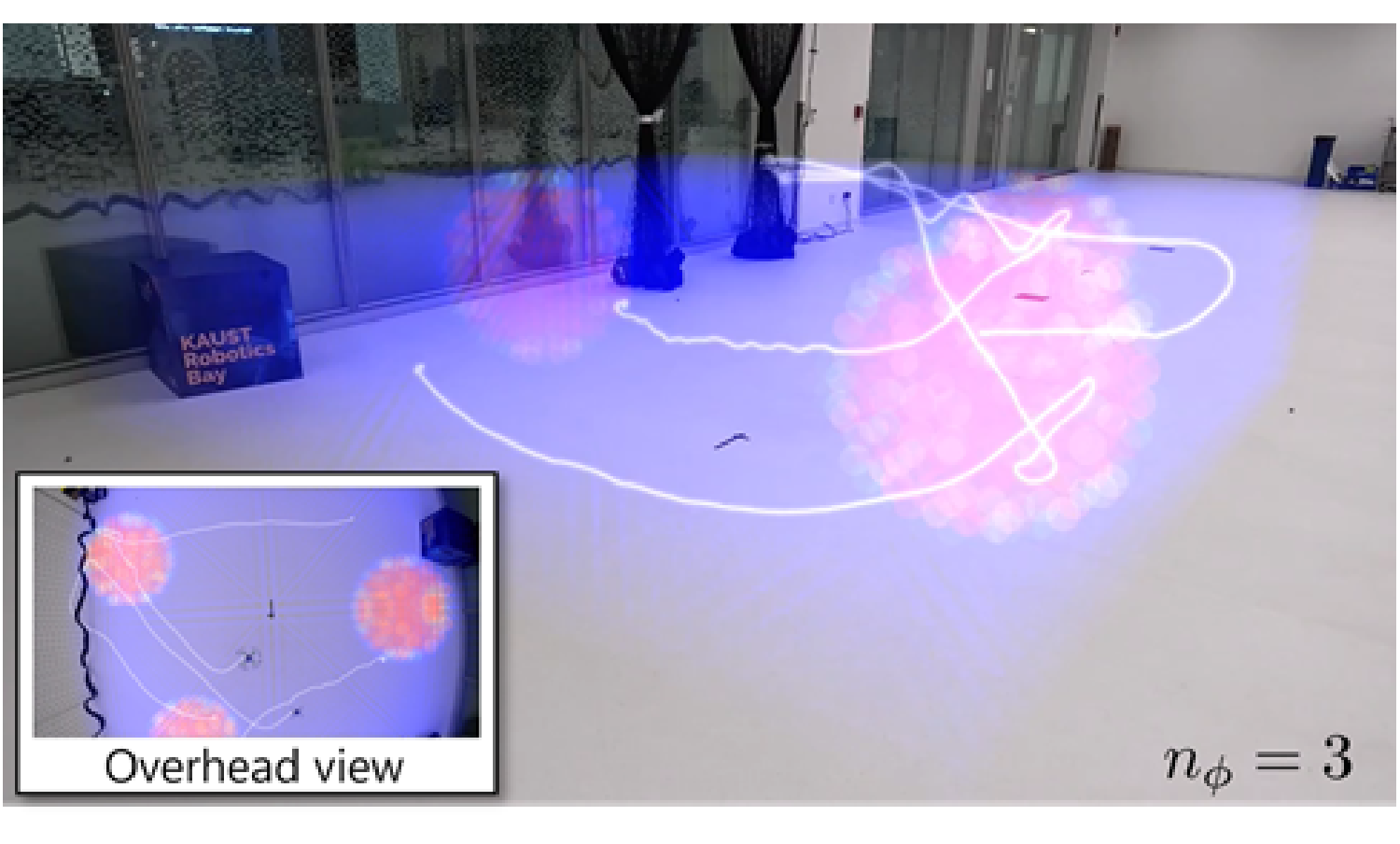}
    \end{subfigure}
    \hfill 
    \begin{subfigure}[b]{0.32\linewidth}
        \includegraphics[width=\textwidth]{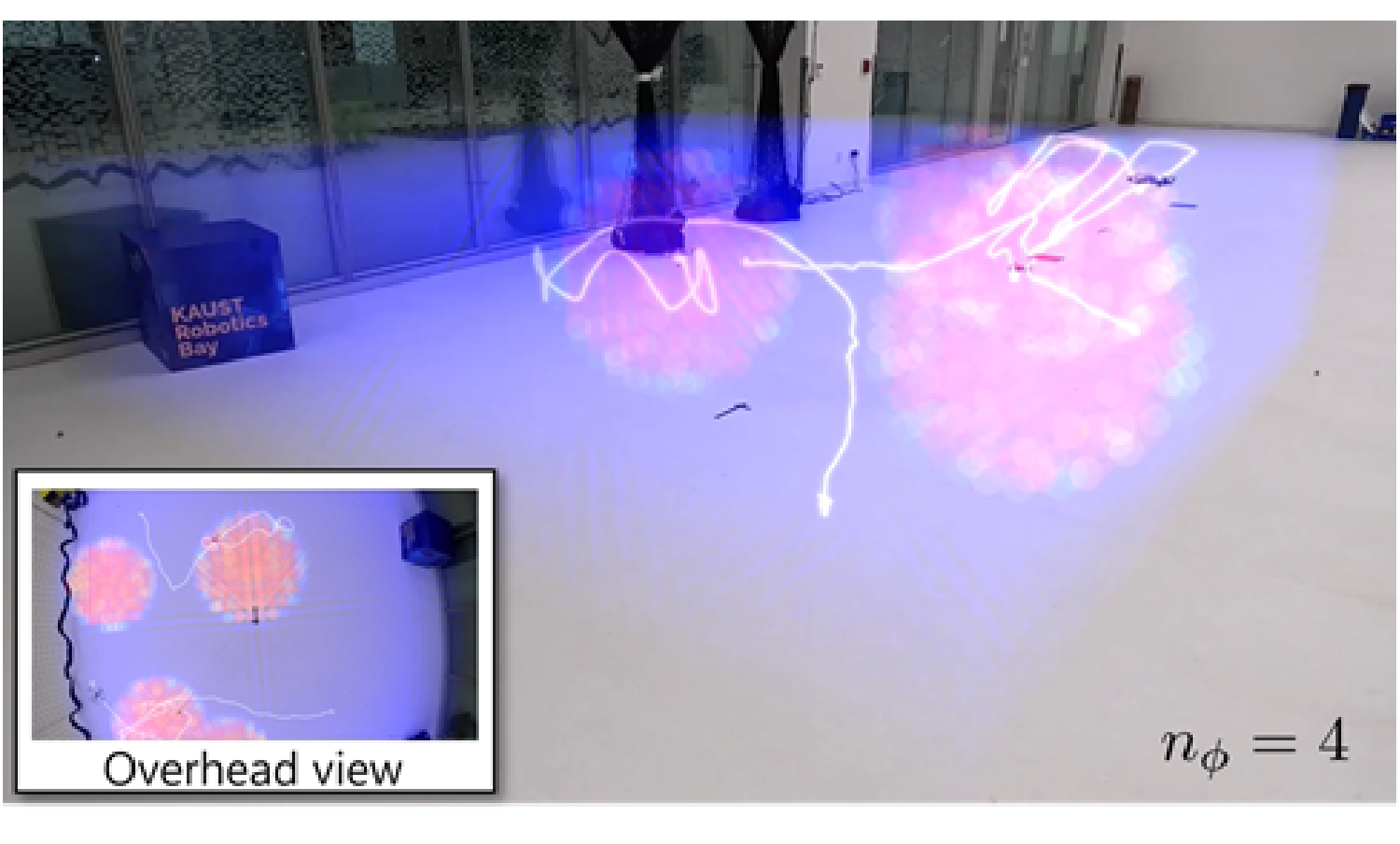}
    \end{subfigure}
    \begin{subfigure}[b]{0.32\linewidth}
        \includegraphics[width=\textwidth]{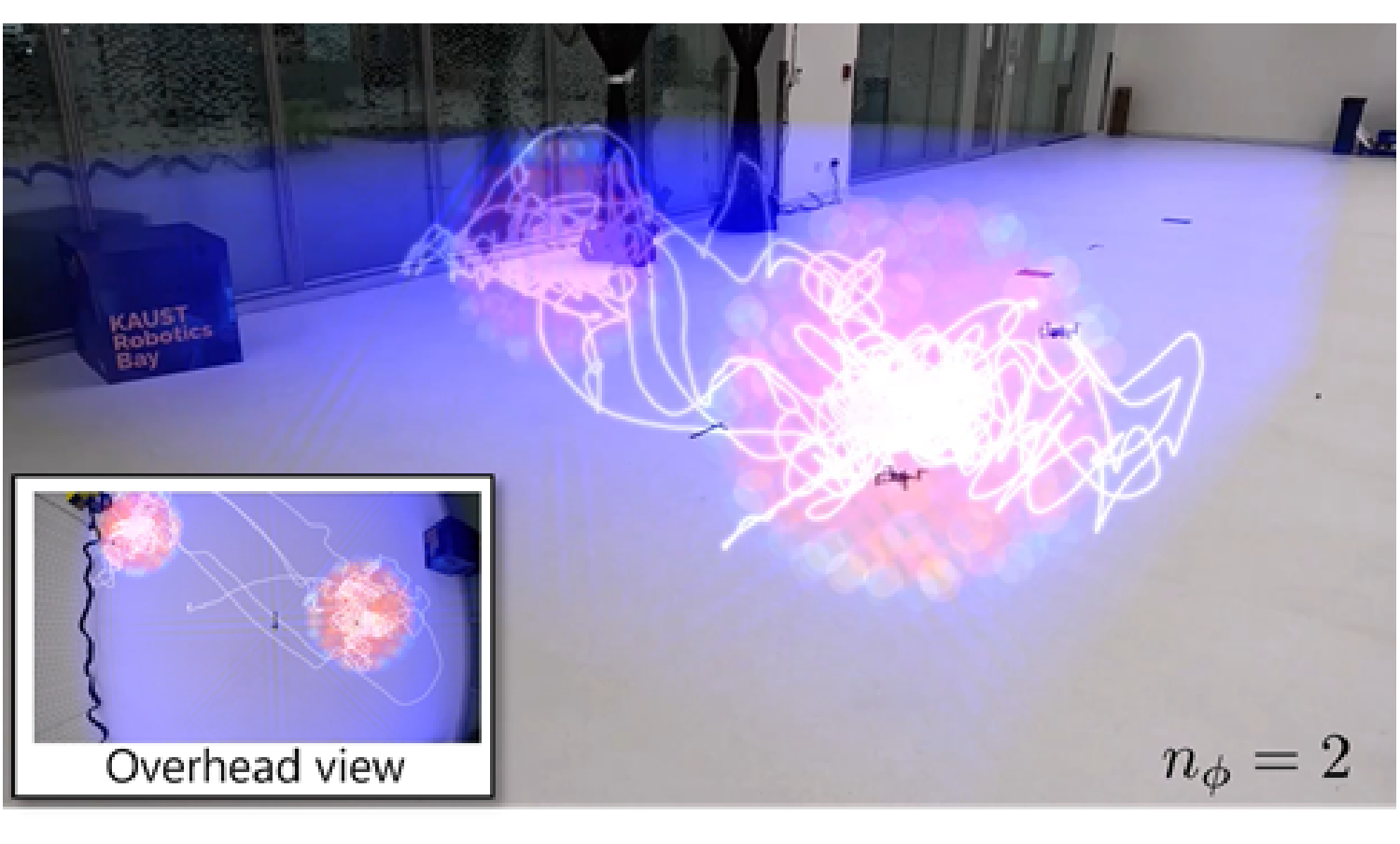}
    \end{subfigure}
    \hfill 
    \begin{subfigure}[b]{0.32\linewidth}
        \includegraphics[width=\textwidth]{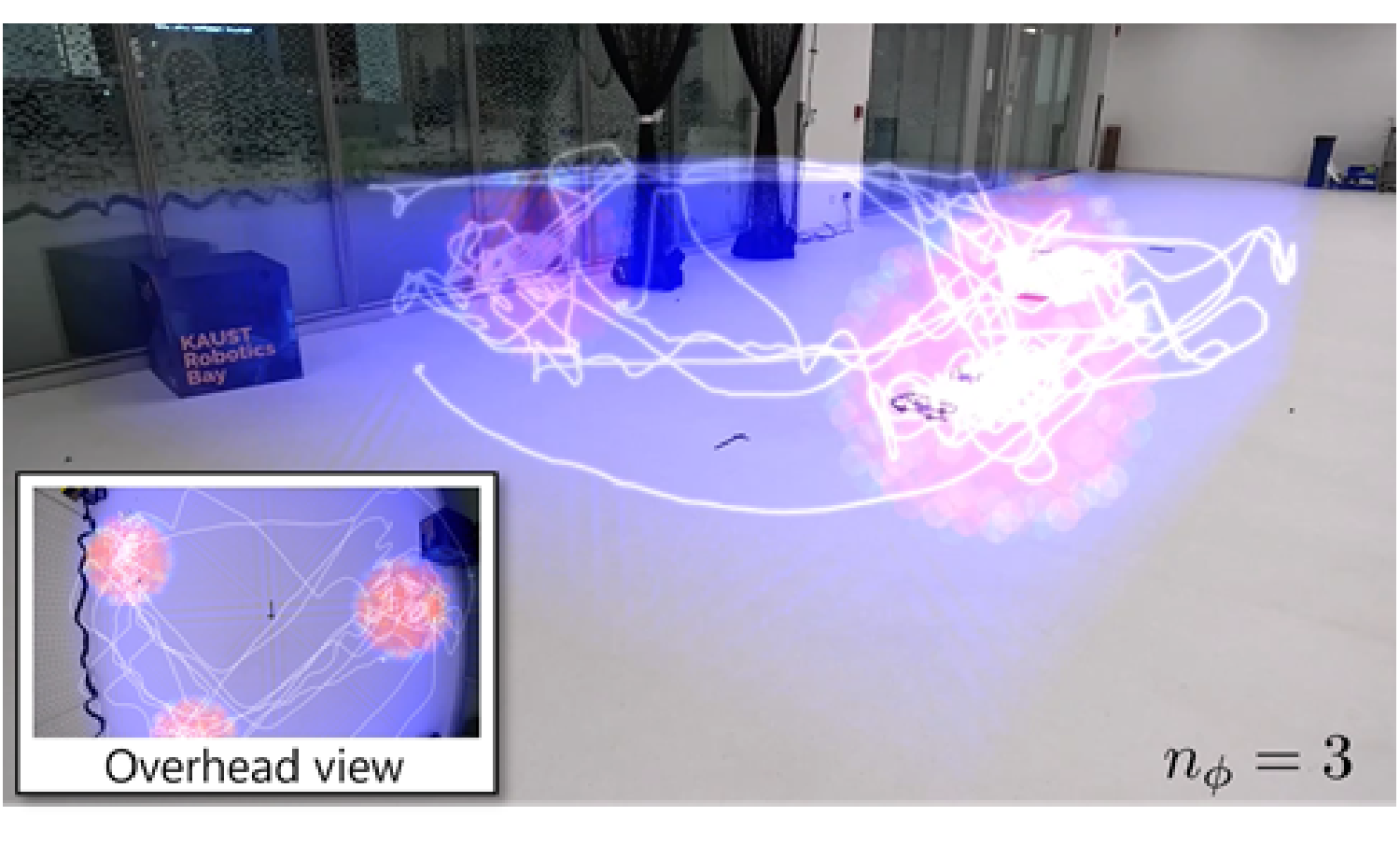}
    \end{subfigure}
    \hfill 
    \begin{subfigure}[b]{0.32\linewidth}
        \includegraphics[width=\textwidth]{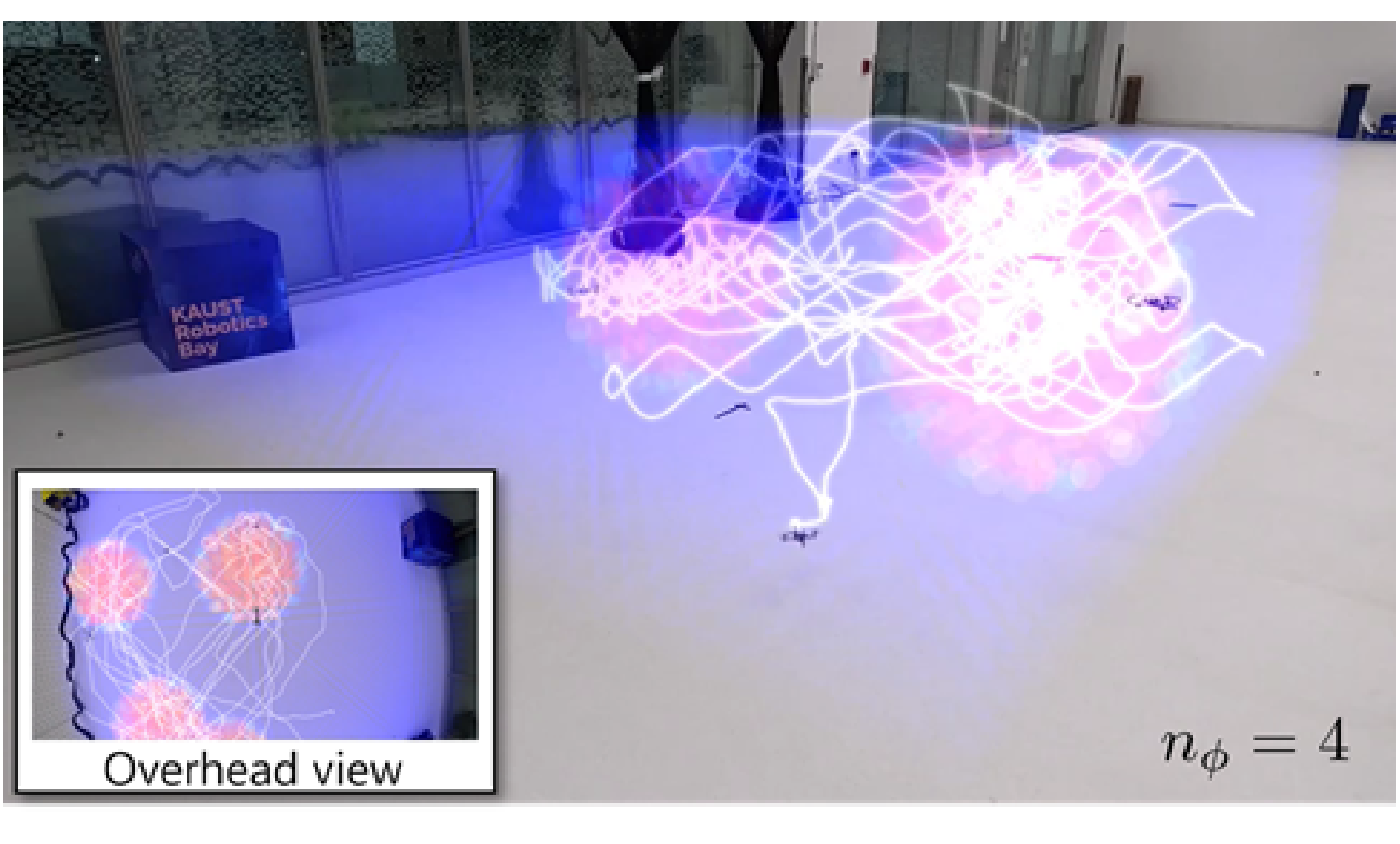}
    \end{subfigure}
    \caption{Snapshots of the three experiments with $n_\phi = 2$, $3$, and $4$ peaks (left to right) at $t = \SI{30}{s}$ (top) and $t = \SI{150}{s}$ (bottom). Target distribution is overlaid in color (red high, blue low) and drone trajectories are shown in white.}
    \label{figs:experiment-ss}
\end{figure*}

We validate the proposed controller on three aerial drones, Crazyflie. A Qualisys motion-capture system provides position measurements, from which $\bar \kappa$ is evaluated every $\SI{0.01}{s}$ on a central computer; the same measurements are fed back to the drones for local control. The central computer solves the QP and transmits $\bm{u}^{\star}_j$ via radio communication to each drone as a velocity reference, which the onboard PID controller tracks.

The flight volume is $[\SI{-2.0}{m}, \SI{2.0}{m}]$ in $x$ and $y$ and $[\SI{0.2}{m}, \SI{1.2}{m}]$ in $z$, with the floor at $z=\SI{0.0}{m}$. We consider the form \eqref{eq:multi-gauss-normalize} with $n_\phi \in \{2, 3, 4\}$ peaks, $\sigma_{\phi} = 0.12$, and peak centers $\bm{q}_{\phi_i}$ drawn at random at least $4\sigma_{\phi}$ away from the boundary. The maximum velocity of the robot is $u_{\max} = \SI{0.2}{m/s}$. The boundary CBF uses $\beta_{\mathrm{b}} (\theta) = 0.9 \, \theta$, chosen so that the QP remains feasible at takeoff from the floor, i.e., outside the safe set. The collision-avoidance CBF uses $r_{\min} = \SI{0.3}{m}$ and $\beta_{\mathrm{c}}(\theta) = 0.25\, \theta$, chosen so that physical contact, which occurs at a separation of about $\SI{0.1}{m}$, is avoided even in the worst case. 
Initial horizontal positions are random, and the initial hovering height is $\SI{0.4}{m}$.
All other controller parameters, the grid spacing, and the update periods are as in the simulations.

\begin{figure}
    \centering
    \includegraphics[width=1.0\linewidth]{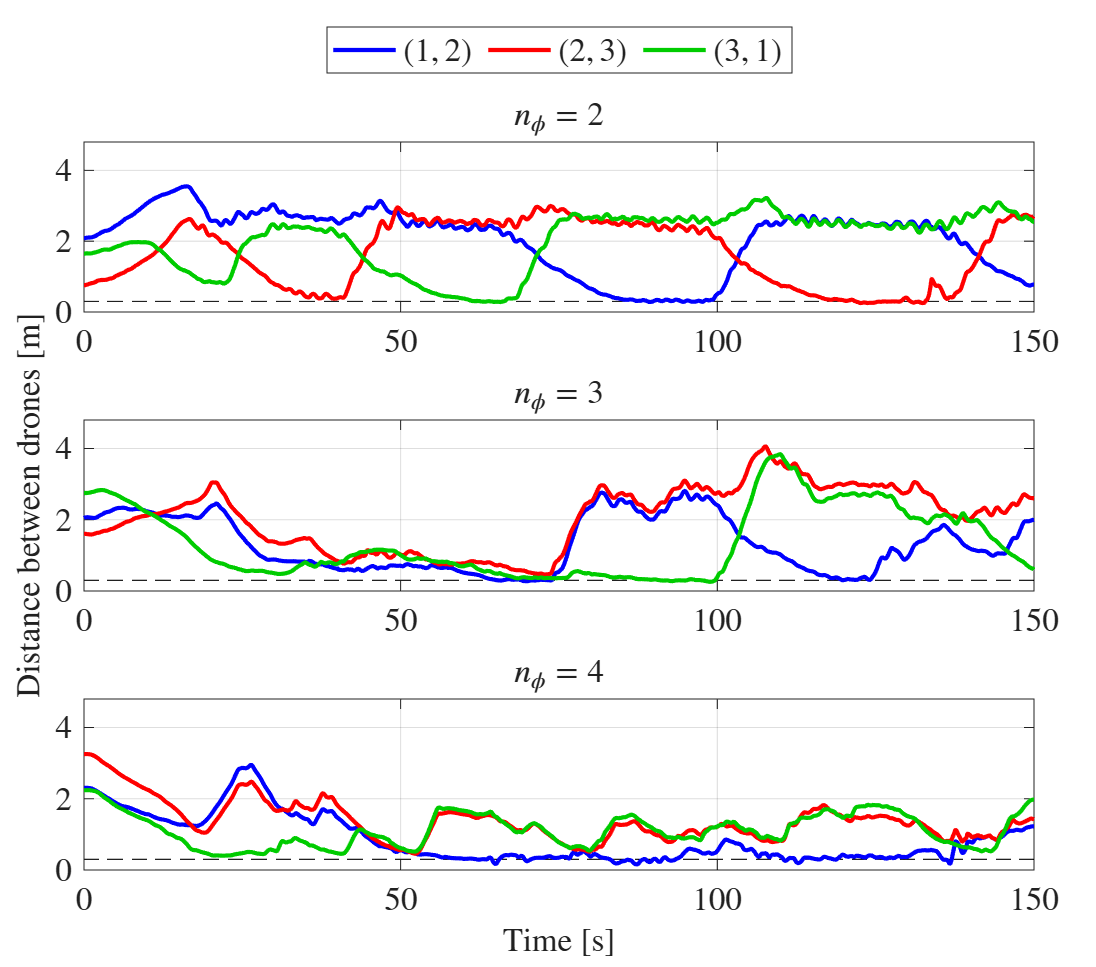}
    \caption{Pairwise distances between drones $\|\bm p_j - \bm p_k \|_2$ for $n_\phi = 2$, $3$, and $4$ (top to bottom). Legend entries indicate the pair $(j,k)$. The dashed line marks $r_{\min} = \SI{0.3}{m}$.}
    \label{fig:exp-distance-drone}
\end{figure}

Figure~\ref{figs:experiment-ss} shows video snapshots from the side and top, overlaid with the target distribution (red high, blue low) and the drone trajectories (white); the full video is provided in supplemental material. Already at $t=\SI{30}{s}$, the drones concentrate on the high-importance regions. The video shows converging on the same peak and drones spreading over different peaks. Trajectories occasionally exceed the boundary because the actual system dynamics have a higher relative degree than the single-integrator model, which causes overshoot; however, as the CBF conditions render the safe set asymptotically stable \cite{robot-ecology}, the drones recover the safe set and the experiment continues without interruption.
Figure~\ref{fig:exp-distance-drone} shows the pairwise distances for each scenario. They remain above $r_{\min}$ (dashed black line) most of the time; the largest violation, again due to the system dynamics model mismatch, reduced the distance to $\SI{0.15}{m}$, but no contact occurred in any experiment. Incorporating higher-order system dynamics into CBF design is left as future work.

\begin{figure}
    \centering
    \includegraphics[width=1.0\linewidth]{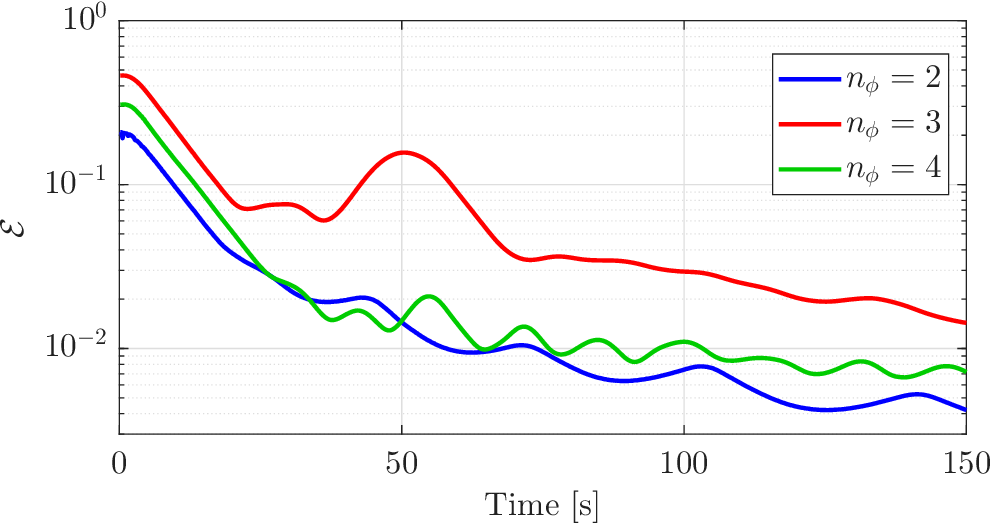}
    \caption{Time evolution of the ergodic metric $\mathcal E$ in \eqref{eq:smooth-ergodic} in the experiments with $n_\phi = 2$, $3$, and $4$ peaks.}
    \label{fig:exp-ergodic-metric}
\end{figure}

Figure~\ref{fig:exp-ergodic-metric} shows the ergodic metric from the time at which $d$ begins to accumulate until $t=\SI{150}{s}$. In all three scenarios, the metric decreases overall.
As in the two-peak simulations, it rises shortly before the drones leave a peak, keeps rising during the transit, and falls once they arrive at the next peak. The pronounced rise between $40\si{s}$ and $50\si{s}$ for $n_\phi=3$ occurred when all three drones gathered at a single peak, where the reduction of the metric was limited, and then moved together toward another peak that had not been sufficiently visited; the metric decreased steadily thereafter. Despite transient constraint violations caused by model mismatch, the proposed controller thus achieves ergodic behavior while maintaining safe operation in real hardware.

We lastly examine the sensitivity to $\sigma_{\max}$.
When no safety constraint is active, the direction $\bm u_j$ is determined by $\bm{b}_j$, which integrates the mismatch $d - \bar\phi$ weighted by $\bm \rho_j$. For a given $\sigma$, $\bm{\rho}_j$ is concentrated within a distance of order $\sigma$ from $\bm p_j$, so with a small $\sigma_{\max}$ the input responds only to mismatch in the immediate vicinity of the robot and the behavior becomes myopic; larger scales let distant mismatch influence the direction. Since $\bm{\rho}_j$ scales with $\frac{1}{\sigma^2}$, the behavior is expected to be most sensitive to $\sigma_{\max}$ when it is small.

\begin{figure}
    \centering
    \includegraphics[width=1.0\linewidth]{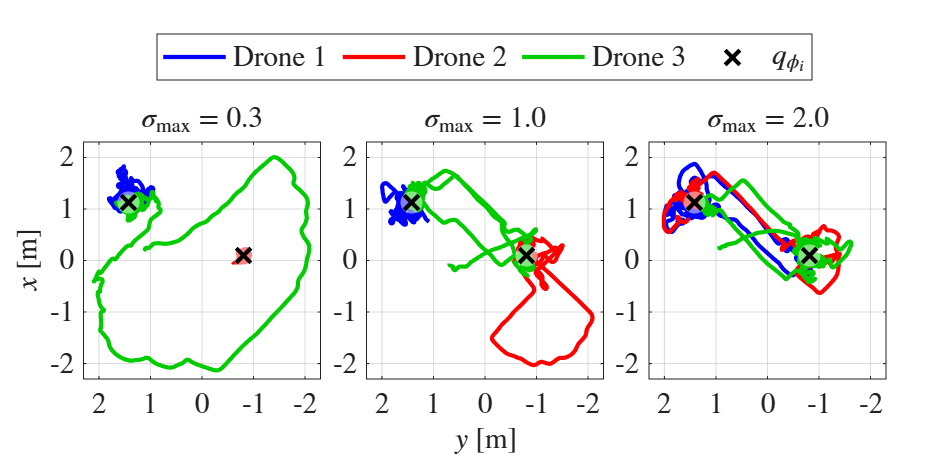}
    \caption{Drone trajectories in the $xy$-plane in the two-peak experiment for $\sigma_{\max} = 0.3$, $1.0$, and \SI{2.0}{m} (left to right). Crosses mark the peak locations.}
    \label{fig:two-peak-sigma-xyposition}
    \vspace{-1.0em}
\end{figure}

To test this, we ran the two-peak experiment with $\sigma_{\max} = 0.3, 1.0, 2.0$, using the same random seed for the peak locations $\bm{q}_{\phi_i}$ and similar initial drone configurations across trials. Figure~\ref{fig:two-peak-sigma-xyposition} shows the resulting trajectories. With $\sigma_{\max} = 0.3$, Drone 3 initially moved along the boundary instead of being drawn to either peak; it eventually reached the peak monitored by Drone 1 and stayed there, while Drones 1 and 2 never switched peaks. With $\sigma_{\max} = 1.0$, Drone 3 transitioned directly between the peaks, although Drone 2 once moved along the boundary and took some time to return to a peak. With $\sigma_{\max} = 2.0$, all drones transitioned directly between the peaks more freely.
The difference between $\sigma_{\max} = 1.0$ and $\sigma_{\max} =2.0$ is smaller than that between $\sigma_{\max} = 0.3$ and $\sigma_{\max} = 1.0$, consistent with the weaker dependence of $\bm{\rho}_j$ on $\sigma$ at larger scales.
Since the cost of evaluating the ergodic metric and its time derivatives grows with $\sigma_{\max}$, we conclude that $\sigma_{\max}$ should be large enough to avoid myopic behavior, beyond which the choice trades computational cost against more direct transitions between peaks.

\section{Conclusion and Future Works}\label{sec:conclusion}
We have presented a single-layer ergodic controller in QP form. A Gaussian-kernel ergodic metric was introduced that, unlike the classical indicator-based metric,  is differentiable in time, which allows the decay of the metric to be imposed as a time-varying CBF constraint alongside hard CBF constrains for region containment and collision avoidance. The simulations showed ergodicity comparable to SMC for a single robot and substantially better ergodicity than a hierarchical SMC-CBF baseline for multiple robots. The experiments with three Crazyflies confirmed ergodic behavior with safe operation on real hardware.

Future work includes distributed controller design, adaptation of time-varying target distributions, extension to higher-order dynamics with sensor models, and  validation in more cluttered environments, all of which are needed for deployment in real environmental monitoring.

\balance

\appendix
\subsection{Proof of Theorem~\ref{theo:no_actuator_limits}}
We utilize similar machinery to~\cite{robot-ecology}. To this end, stack the $q = m + \binom{m}{2}$ constraints~\eqref{eq:qp-constraint-ergodic-region-obj-const-region}-\eqref{eq:qp-constraint-ergodic-region-obj-const-col}
\begin{align*}
         N\bm{u}&+\bm{\beta}\ \ge\ 0,
    \qquad \bm{\beta}:=C\bm{h}_{\mathrm{sf}} \\
    \bm{h}_{\mathrm{sf}}&:=\big(
    h_{\mathrm{b},1},\dots,h_{\mathrm{b},m},\,
    h_{\mathrm{c},12},\dots,h_{\mathrm{c},(m-1)m}\big)\in\mathbb{R}^{q},
    \\
    N&:=\frac{\partial\bm{h}_{\mathrm{sf}}}{\partial\bm{p}}\in\mathbb{R}^{q\times 3m}, \quad C:=\mathrm{diag}\big(c_{\mathrm{b}}I_m,\,c_{\mathrm{c}}I_{\binom{m}{2}}\big)\succ0
\end{align*}
Define the  multipliers $\lambda\ge0$ for ergodicity constraint,
$\bm{\nu}\in\mathbb{R}^{q}_{\ge0}$ for safety constraints and $\mu\ge0$ for
the slack, the KKT conditions read (dropping arguments for clarity)
\begin{subequations}\label{eq:kkt}
\renewcommand{\theparentequation}{KKT}
\renewcommand{\theequation}{\theparentequation-\arabic{equation}}
\begin{align}
    &2\bm{u}^{\star}+\lambda\bm{b}-N^{\top}\bm{\nu}=0,
    \label{eq:kkt-u}\\
    &2w\delta^{\star}-\lambda-\mu=0,
    \label{eq:kkt-delta}\\
    &\psi-\bm{b}^{\top}\bm{u}^{\star}+\delta^{\star}\ge0,\quad
    N\bm{u}^{\star}+\bm{\beta}\ge0,\
    \delta^{\star}\ge0,
    \label{eq:kkt-primal}\\
    &\lambda\ge0,\quad \bm{\nu}\ge0,\quad \mu\ge0,
    \label{eq:kkt-dual}\\
    &\lambda\big(\psi-\bm{b}^{\top}\bm{u}^{\star}+\delta^{\star}\big)=0, \nonumber\\
     &\operatorname{diag}(\bm{\nu})\big(N\bm{u}^{\star}+\bm{\beta}\big)=0,\ \
     \mu\delta^{\star}=0,
    \label{eq:kkt-comp}
\end{align}
\end{subequations}
It is easy to see that $\mu =0$ (by a contradiction argument), and from~\eqref{eq:kkt-delta},
\begin{equation*}\label{eq:slack-multiplier}
    \delta^{\star}=\cfrac{\lambda}{2w}\ \ge\ 0 ,
\end{equation*}
so \eqref{eq:qp-constraint-ergodic-region-obj-const-slack} is redundant. Let $\mathcal{A}$ index the rows with positive multipliers, let $N_{\mathcal{A}}$
and $\bm{\beta}_{\mathcal{A}}$ collect the corresponding rows of $N$ and entries
of $\bm{\beta}$.
Multiplying \eqref{eq:kkt-u} by $N_{\mathcal{A}}$, and using
$N_{\mathcal{A}}\bm{u}^{\star}=-\bm{\beta}_{\mathcal{A}}$ from
\eqref{eq:kkt-comp}, isolates
$\bm{\nu}_{\mathcal{A}}=(N_{\mathcal{A}}N_{\mathcal{A}}^{\top})^{\#}
(\lambda N_{\mathcal{A}}\bm{b}-2\bm{\beta}_{\mathcal{A}})$, with $(\cdot)^{\#}$ the Moore--Penrose inverse. Consequently, with
\eqref{eq:kkt-u} one has
\begin{align*}
    \bm u^\star &= -\tfrac{\lambda^{\star}}{2}P^{\perp}\bm{b}
    - \bm v
\end{align*}
where herein we defined $\bm{v} =N_{\mathcal{A}}^{\top}\big(N_{\mathcal{A}}N_{\mathcal{A}}^{\top}\big)^{\#}
\bm{\beta}_{\mathcal{A}}$, and $P^\perp = I_{3m}-N_\mathcal{A}^\top(N_\mathcal{A}N_\mathcal{A}^\top)^\#N_\mathcal{A}$ the orthogonal projector onto $\operatorname{ker}N_{\mathcal{A}} $. If $\lambda>0$, and by~\eqref{eq:kkt-comp} and $\bm{b}^{\top}P^{\perp}\bm{b}=\|P^{\perp}\bm{b}\|_2^2$ one has for $\tilde\psi :=\psi+\bm{b}^{\top}\bm v$ that
\begin{align*}
    0 & =\psi-\bm{b}^{\top}\bm{u}^{\star}+\tfrac{\lambda}{2w}
     =\tilde\psi+\tfrac{\lambda}{2}\Big(\|P^{\perp}\bm{b}\|_2^2+\tfrac1w\Big).
\end{align*}
If instead $\lambda=0$, then $\delta^{\star}=0$, $\bm{u}^{\star}=-\bm{v}$ and
\eqref{eq:kkt-primal} gives $\tilde\psi\ge0$, so, 
\begin{align*}
         \lambda^{\star}
    &= \max\left\{0,\frac{-2\,\tilde\psi}{\|P^{\perp}\bm{b}\|_2^2+w^{-1}}\right\} .
\end{align*}
Now note that whenever $\bm{h}_{\mathrm{sf}}\ge0$, the pair $(\bm{u},\delta)=(\bm{0}, \cdot)$
is feasible, since $N\bm{0}+\bm{\beta}=C\bm{h}_{\mathrm{sf}}\ge0$. Also while the
QP is feasible its solution gives
$\dot{\bm{h}}_{\mathrm{sf}}+C\bm{h}_{\mathrm{sf}}\ge0$, so that multiplying by
$e^{Ct}$ and integrating on $[t_0,\ t]$ yields
\begin{equation*}\label{eq:safety-lower}
    \bm{h}_{\mathrm{sf}}(t)\ \ge\ e^{-C(t-t_0)}\bm{h}_{\mathrm{sf}}(t_0)\ \ge\ 0 .
\end{equation*}
Safety thus reproduces feasibility and by assumption safety holds at
$t_0$, a standard continuity argument gives feasibility of
\eqref{eq:qp-constraint-ergodic-region} and the above inequality
$\forall t\ge t_0$. The QP solution satisfies $\dot h+c_\alpha h\ge-\delta^{\star}$. Multiplying by $e^{c_\alpha t}>0$ and integrating on $[t_0,t]$,
\begin{equation*}
    h(t)\ \ge\ h(t_0)e^{-c_\alpha(t-t_0)}-\delta_c(t),
\end{equation*}
Since $-h(t_0)\le\eta$, substituting
the above into~\eqref{eq:def-h} isolating $\dot{\mathcal{E}}$ gives
\begin{equation*}
    \dot{\mathcal{E}}\ \le\ -\gamma\mathcal{E}
    +\eta e^{-c_\alpha(t-t_0)}+\delta_c(t).
\end{equation*}
Multiplying by $e^{\gamma(t-t_0)}>0$, integrating on $[t_0,t]$ with
$\int_{0}^{(t-t_0)}e^{(\gamma-c_\alpha)r}dr
=\big(e^{(\gamma-c_\alpha)(t-t_0)}-1\big)/(\gamma-c_\alpha)$, and multiplying by
$e^{-\gamma(t-t_0)}$ gives
\begin{align*}
    \mathcal{E}(t)\; & \le\; \mathcal{E}(t_0)e^{-\gamma(t-t_0)}
    +\frac{\eta}{\gamma-c_\alpha}\Big(e^{-c_\alpha(t-t_0)}-e^{-\gamma(t-t_0)}\Big)
    \\ \quad &+\int_{t_0}^{t}e^{-\gamma(t-\tau)}\delta_c(\tau)\,d\tau .
\end{align*}
The second term is nonnegative and bounding it by
$\tfrac{\eta}{|\gamma-c_\alpha|}e^{-\min\{\gamma,c_\alpha\}(t-t_0)}$ yields
\eqref{eq:ergodic_convergence_rate}.

\end{document}